\documentclass[10pt, conference]{ieeeconf} 
\IEEEoverridecommandlockouts
\usepackage{cite}
\usepackage{amsmath,amssymb,amsfonts}
\usepackage{algorithmic}
\usepackage{graphicx}
\usepackage{textcomp}
\usepackage{xcolor}

\usepackage{url}

\usepackage{cleveref}
\usepackage{flushend}

\def\BibTeX{{\rm B\kern-.05em{\sc i\kern-.025em b}\kern-.08em
    T\kern-.1667em\lower.7ex\hbox{E}\kern-.125emX}}

\DeclareMathAlphabet{\mathcal}{OMS}{cmsy}{m}{n}
\graphicspath{{images/}}

\usepackage[]{caption}
\renewcommand{\thetable}{\arabic{table}}

\begin{document}

\title{\LARGE \bf Relation-based Motion Prediction using Traffic Scene Graphs}

\author{
Maximilian~Zipfl$^{1}$, 
Felix Hertlein$^{1}$, 
Achim~Rettinger$^{2}$, 
Steffen Thoma$^{1}$, 
Lavdim Halilaj$^{3}$, \\
Juergen Luettin$^{3}$, 
Stefan Schmid$^{3}$, 
Cory Henson$^{4}$
\thanks{$^{1}$FZI Research Center for Information Technology, Karlsruhe, Germany
{\tt\small \{zipfl, hertlein, thoma\}@fzi.de}}%
\thanks{$^{2}$Trier University, Trier, Germany
{\tt\small rettinger@uni-trier.de}}%
\thanks{$^{3}$Bosch Corporate Research, Renningen, Germany
{\tt\small \{lavdim.halilaj, juergen.luettin, stefan.schmid\}@de.bosch.com}}%
\thanks{$^{3}$Bosch Research and Technology Center, Pittsburgh, PA, USA
{\tt\small cory.henson@us.bosch.com}}%
}%

\maketitle

\begin{abstract}
Representing relevant information of a traffic scene and understanding its environment is crucial for the success of autonomous driving.
Modeling the surrounding of an autonomous car using semantic relations, i.e., how different traffic participants relate in the context of traffic rule based behaviors, is hardly been considered in previous work. 
This stems from the fact that these relations are hard to extract from real-world traffic scenes. 
In this work, we model traffic scenes in a form of spatial semantic scene graphs for various different predictions about the traffic participants, e.g., acceleration and deceleration. 
Our learning and inference approach uses Graph Neural Networks (GNNs) and shows that incorporating explicit information about the spatial semantic relations between traffic participants improves the predicdtion results. 
Specifically, the acceleration prediction of traffic participants is improved by up to 12\% compared to the baselines, which do not exploit this explicit information.
Furthermore, by including additional information about previous scenes, we achieve 73\% improvements.
\end{abstract}

\section{Introduction}

In road traffic, there exist many influencing factors important to safely drive from one point to another. 
While a number of these factors can be considered static, for example the road infrastructure, the behavior of the various traffic participants changes dynamically over time. 
Therefore, the behavior of nearby traffic participants is important in determining the appropriate action of an autonomous vehicle, e.g. steering or braking. 
Traffic participants might react differently depending on their particular situation, so a model about their intentions, indicating whether they will brake soon, would provide added value. 
To obtain that, in addition to the position of traffic participants, knowledge about how they relate to other traffic participants is crucial. 
This is particularly relevant for autonomous driving when the autonomous agent needs to detect and evaluate the current traffic situation (scene) in order to generate a trajectory. Human drivers derive an understanding of the underlying traffic situation directly from their perceived environment and their knowledge about traffic rules, as well as experience from previous social interactions and behaviors. In this work, we incorporate explicit spatial information about relations among the various traffic participants, and clearly demonstrate the importance of this information. For autonomous agents, it is desirable to have an explicit description of the road environment or driving context in order to evaluate the influences of individual factors and ensure the safety requirements of the autonomous vehicle. To this end, we use semantic scene graphs as a way to describe the driving context of an autonomous agent, including the relation to nearby traffic participants independently of scene types (constellation of traffic participants) or road geometry. This description is represented in a form that can be used as contextual information for the motion planning of an autonomous agent.

This paper is structured as follows: In \Cref{sec:sota} we provide a review of existing traffic scenario descriptions and related work in the context of motion prediction.
\Cref{sec:scene_description} describes a short overlook over the used scene representation model and how certain graph's attributes are derived. 
In \Cref{sec:implementation}, our implementation in regard to the used datasets is discussed. Moreover, two net architectures are proposed on how to extract information from a scene graph. The results of the models are shown and evaluated in \Cref{sec:experiments}.
Finally, in \Cref{sec:conclusion} we conclude this contribution.

\section{State of the Art}
\label{sec:sota}
\subsection{Scene and Scenario Descriptions}

In recent years, knowledge graphs have found their way into a wide variety of domains. They are used to store knowledge in a structured and extensible graph representation and enable agents to query them for all kinds of information. 
Ulbrich et al.~\cite{Ulbrich2014GraphbasedCR} present an approach for representing and modeling context and environment in driving scenarios. 
It comprises various layers for describing information about lane, traffic rules, participants, and the overall situation.
Henson et al.~\cite{DBLP:conf/semweb/HensonSTK19} describes a semantic model with the most important concepts in a driving scene such as \emph{sequence}, \emph{scene}, \emph{participant} and their inter-relationships.
A knowledge-graph based approach for representing and fusing heterogeneous types of information of traffic scenes is presented in~\cite{Halilaj2021AKnowledge}.
The integrated knowledge is then used along with graph neural networks for classification of driving situations.

\subsection{Motion Prediction}
A general survey about motion prediction for automated driving can be found in~\cite{Leon2019ARO}.
A prominent method that reasons jointly about the 3D scene layout of intersections as well as the location and orientation of objects in the scene is proposed in~\cite{Geiger20143DTS}. 
An approach that includes high-level semantic information in a spatial grid, combined with CNN to model complex scene context for trajectory prediction, is described in~\cite{Hong2019RulesOT}.
Casas et al.~\cite{Casas2018IntentNetLT} present IntentNet, a one-stage detector and forecaster based on 3D point clouds of a LiDAR sensor and dynamic maps of the environment. 
MutliPath~\cite{Chai2019MultiPathMP} uses state-sequence anchors that correspond to model of trajectory distributions. 
At inference, the model predicts a discrete distribution over the anchors and regresses offsets from anchor waypoints along with uncertainties. 
These yield a Gaussian mixture at each time step.
Zhao et al.~\cite{Zhao2020TNTTT} present the Target-driveN Trajectory (TNT) framework, that consists of three steps: predict the agent's potential target states into the future, generate trajectory state sequences conditioned on targets, estimate trajectory likelihoods and final compact trajectory predictions.

\subsection{Graph-based Motion Predictions}
A survey of deep learning-based vehicle behavior prediction for automated driving is provided by~\cite{Mozaffari2019DeepLV}.
Diel et al.~\cite{Diehl2019GraphNN} compare the trajectory prediction performance of Graph Convolutional Networks (GCN)~\cite{Kipf2017SemiSupervisedCW} with Graph Attention Networks (GAT)~\cite{DBLP:conf/iclr/VelickovicCCRLB18} and propose modifications to the task of vehicle behavior prediction.
An attention-based spatio-temporal Graph Neural Networks (GNN)~\cite{Scarselli2009TheGN} for pedestrian trajectory prediction is proposed in~\cite{Zhou2021ASTGNNAA,Yu2020SpatioTemporalGT,Haddad2019SituationAwarePT}.
Li et al.~\cite{Li2019GRIPGI} present a graph based interaction-aware trajectory prediction (GRIP) approach.
This models the interaction between the vehicles using a GCN and graph operations.
Mo et al.~\cite{Mo2021GraphAR} describe a GNN-RNN based approach for trajectory prediction, where the vehicles are modelled by an RNN and the interactions between them as a directed graph using a GNN.
The output of the model is further processed by a Long Short Term Memory (LSTM)~\cite{hochreiter_long_1997}.
An enhanced version~\cite{Li2019GRIPEG} uses both fixed and dynamic graphs for trajectory prediction.
Frenet coordinate systems are well investigated for motion planning and trajectory generation~\cite{Werling2010OptimalTG}.
The Frenet representation is adopted for multi-agent interaction aware trajectory prediction by a GNN-based solution \cite{ma_multi-agent_2021}.
Instead of a Cartesian coordinate system and global context images, it uses the pairwise relations between agents and pairwise context information.
Fang et al.~\cite{Fang2019OntologybasedRA} propose an approach for long-term behavior prediction using ontology-based reasoning that considers interactions between traffic participants.
Their likelihood behavior is inferred by a Markov logic network.
Gao et al.~\cite{Gao2020VectorNetEH} presented VectorNet, a vectorized definition of the scene where unified representations are learned from their vectorized form.
The graphic extent of a road feature can be a point, a polygon, or a curve in geographic coordinates.
A GNN is then used to incorporate the set of vectors, where each vector is treated as a node in the graph.
Li et al.~\cite{Li2021SpatioTemporalGD} present a spatio-temporal graph dual-attention network for multi-agent prediction and tracking. 
It incorporates relational inductive biases, a kinematic constraint layer
and leverages both trajectory and scene context information.

\section{Scene Representation}
\label{sec:scene_description}
In this chapter, we describe the procedure for creating the scene description for a concrete application example.
To analyze realistic traffic scenes and the resulting behaviors, we use openly available datasets, which we describe in detail in \Cref{sec:dataset}.

As the foundation for representing relations between traffic participants and describing the underlying traffic scenes, we apply the \emph{Semantic Scene Model} (SSM), which is based on our previous work \cite{zipfl2021traffic}. 
This model establishes relations between the traffic participants in the scene based on the given road topology. 
The traffic scene is mapped to a graph, in which the traffic participants are described by nodes and their relations by edges.

\begin{figure}[htbp]
  \centering
  \includegraphics[width=0.8\columnwidth]{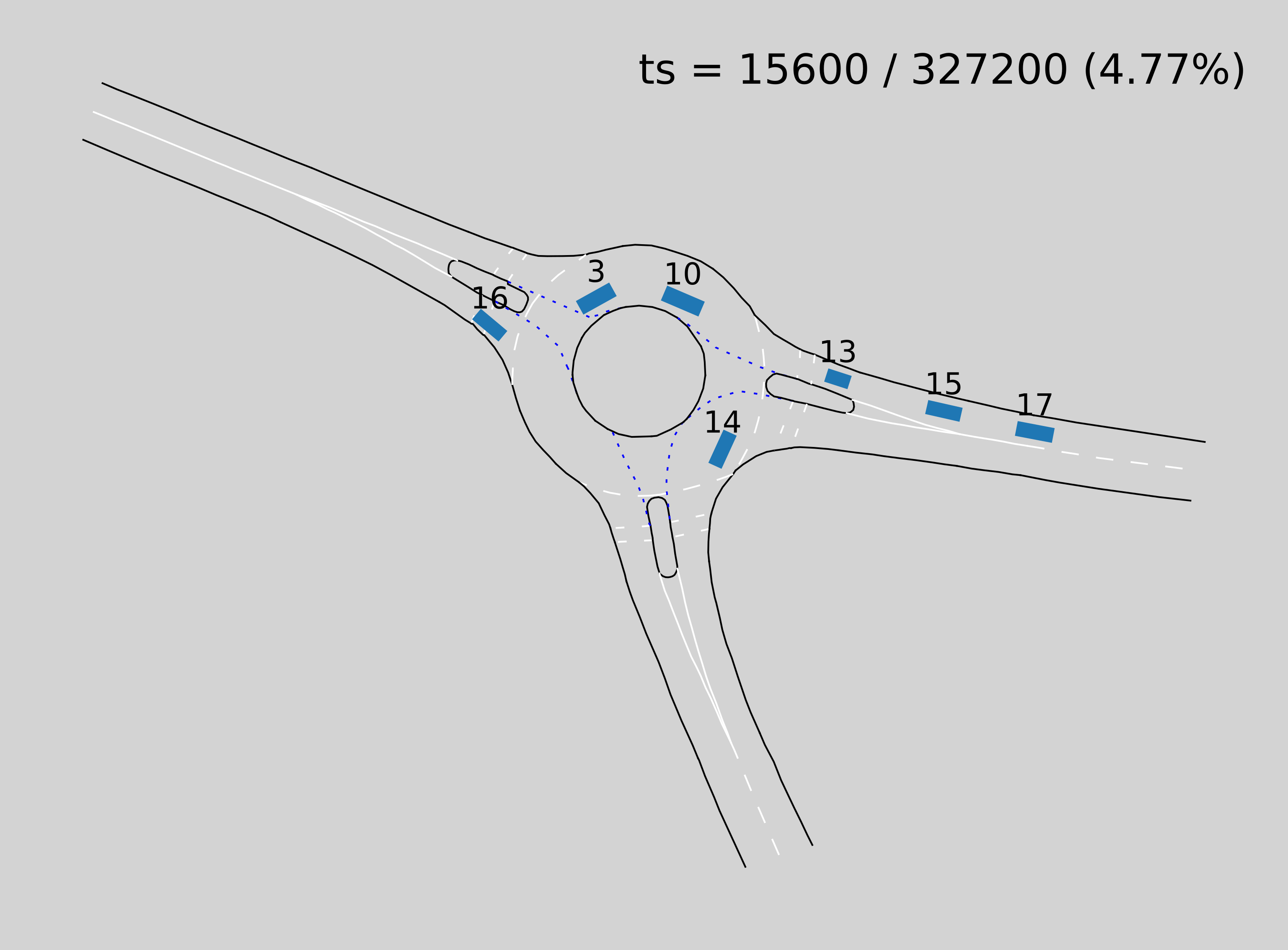}
  \caption{Traffic scene seen from bird's eye view.}
  \label{fig:interaction_roundabout_example}
\end{figure}

\begin{figure}[htbp]
  \centering
  \def\svgwidth{0.7\columnwidth}
  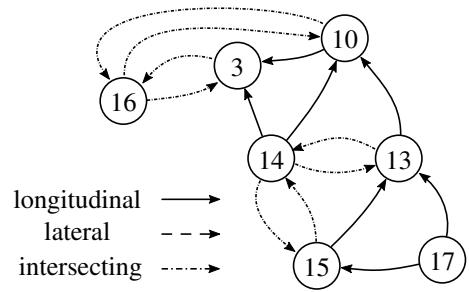
  \caption{Scene graph topology of the traffic scene of \Cref{fig:interaction_roundabout_example}.}
  \label{fig:graph_interaction_roundabout_example}
\end{figure}
\Cref{fig:interaction_roundabout_example} illustrates an exemplary traffic scene from the INTERACTION dataset \cite{interactiondataset} and its corresponding topology of the derived scene graph is depicted in \Cref{fig:graph_interaction_roundabout_example}. 
Each of the seven traffic participants is represented as a node. 
Each node $v$ contains information about its classification (car, pedestrian, truck, ...) and its velocity. Edges between the nodes are of three different types: longitudinal, lateral and intersecting.
This can be seen for example at the nodes or vehicles 10, 13, 14 and 15. If we consider the road layout in \Cref{fig:interaction_roundabout_example}, vehicles 13 and 15 follow vehicle 10. Vehicle 14 also follows vehicle 10, but at the same time, vehicle 13 and  vehicle 15 have an intersecting relation with vehicle 14, since they drive on two roads that will merge eventually. This situation can be illustrated well by the generic graph description (\Cref{fig:graph_interaction_roundabout_example}).

When converting the scene into the scene graph, the individual traffic participants are assigned to lanes. This allows the traffic scene to be viewed in the \emph{Frenet} space, with the roads being the curves.
In addition to the classification of the edge type, its probability and the distance along the roadway, between the two respective entities, is also included as an edge attribute (see \cref{tab:distance_table}). 
When assigning every entity to a lane, the distance to the centerline of the lane and the angular difference between the entity's pose and the lane is calculated. These two values can be used to calculate the assignment probability $P(i)$ of an entity $i$ to a lane.
We define the probability $P(e_{ij})$ of an edge $e_{ij}$ as follows:
\begin{align}
    P(e_{ij}) = P(i) \cdot P(j).
\end{align}
A more detailed overview of the calculation of the probability of each entity, especially with respect to the assignment to multiple lanes, is given in our previous work~\cite{zipfl2021traffic}.

\begin{figure}[htbp]
  \centering
  \def\svgwidth{0.9\columnwidth}
  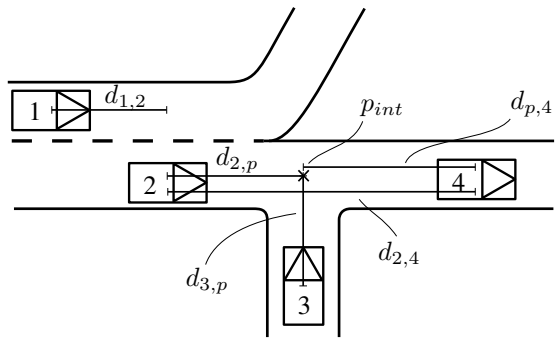 \\

  \caption{Schematic traffic with distances between traffic participants}
  \label{fig:distance_scheme}
\end{figure}

An example of the distance attributes, a schematic traffic scene and the corresponding edges is presented in \Cref{fig:distance_scheme}. 
If the relation between two entities is longitudinal (\texttt{lon}), the distance along the center line of the road to the projection of the entity is calculated, so that a lateral shift of the entity does not affect the distance (see $d_{2,4}$). The principle remains the same for merging lanes. Here, the distance along the road is also determined. The point where the two roads merge is called $p_{int}$. This means that the distance between vehicle (3) and vehicle (4) is noted as $d_{3,p}+d_{p,4}$. For two parallel roads, the distance between the two entities is determined similarly to longitudinal calculation. In this example, vehicle (2) is projected to the same lane as vehicle (1). Then the distance between both traffic participants is measured (compare $d_{1,2}$). For intersecting entities (\texttt{int}), only the distance from the tail-entity to the intersection point (here $p_{int}$) is considered (see $d_{2,p}, d_{3,p}$). 

\begin{table}[htbp]
    \centering
    \begin{tabular}{lccc}
    \hline
    edge  & edge probability & classification & distance \\ \hline
    $e_{12}=(1,2)$ & $P(e_{12})$ & \texttt{lat} & $d_{1,2}$ \\
    $e_{21}=(2,1)$ & $P(e_{21})$ & \texttt{lat} & $-d_{1,2}$ \\
    $e_{23}=(2,3)$ & $P(e_{23})$ & \texttt{int} & $d_{2,p}$ \\
    $e_{32}=(3,2)$ & $P(e_{32})$& \texttt{int} & $d_{3,p}$ \\
    $e_{24}=(2,4)$ & $P(e_{24})$ & \texttt{lon} & $d_{2,4}$ \\
    $e_{34}=(3,4)$ & $P(e_{34})$& \texttt{lon} & $d_{3,p}+d_{p,4}$\\\hline
    \vspace{0.5cm}
  \end{tabular}
    \caption{Structure of the edge attributes of the scene described in \Cref{fig:distance_scheme}.}
    \label{tab:distance_table}
\end{table}

\section{Implementation and Training}
\label{sec:implementation}

\subsection{Dataset}
\label{sec:dataset}
Two different datasets are investigated: first, the PandaSet dataset \cite{xiao2021pandaset}, which provides short sequences (8s). 
Driving scenes are captured by a car equipped with multiple cameras and sensors. 
The captured scenes are divided into single images that depict the state of a scene at a specific point in time.
The dataset comprises complex scenarios typically happening in urban areas like dense traffic and construction sites, as well as a variety of times of day and lighting conditions.

Secondly, the INTERACTION dataset \cite{interactiondataset}, which is recorded by a drone at intersections and roundabouts.
This dataset focuses on the behavior of vehicles on roads in different countries. In addition to the object tracks, the corresponding roads are provided as HD\footnote{high-definition} maps.
Both datasets are recorded at a frequency of $10\,Hz$.

\subsection{Preprocessing}
\label{sec:preprocessing}
We initially convert both datasets to the same data format.
The used output data format is the same as in the INTERACTION dataset. 
Here, all tracks of each traffic participant for each point in time are stored in a csv\footnote{comma-separated values}-file. 
Each state of each traffic participant is uniquely defined by the track-id and the timestamp. In addition, the pose, the classification $\{Car, Pedestrian, Truck,...\}$, the velocity vector and, in the PandaSet, the motion-state $\{Parked, Stopped, Driving\}$ are stored.

The focus of the PandaSet dataset is on object detection using Lidar sensors. Nevertheless, object lists with the individual tracked entities with their pose and classification can be read out. An important input variable is the velocity of the individual traffic participants. In the PandaSet dataset, this is calculated retrospectively based on the change in position over time, followed by a low pass filter to compensate measurement inaccuracies. The original dataset contains partially duplicate annotations for a given entity at a given timestamp. In order to remove the duplicates, we represent each entity by a rotated rectangular polygon and calculate \emph{intersection over union} (IoU) between possible duplicates. 
Starting from an IoU of 0.2 and above, we consider those two to as conflicting. To identify the set of duplicates, we interpret conflicts as edges and involved entities as nodes in a conflict graph $G_{conf}$. 
$G_{conf}$ is used to search for a maximum independent set of nodes such that all nodes are not connected, and delete all other entities.
Furthermore, we removed all vehicles farther away than 80 meters from the ego vehicle since the measuring inaccuracy for far vehicles lead to extreme velocities and accelerations.

\subsection{Net Architecture}
For modelling the traffic scenes, we use a graph representation $G = (V,E)$ on which we can leverage a message passing approach. 
This allows for displaying specific traffic participants and their relation in a machine-readable format.
For further calculations, the graph is described by three data blocks. The node attributes are described in a $n\times\left|v\right|$ matrix, where $n$ describes the number of nodes in the graph $G$ and $\left|v\right|$ the number of attributes of a node $v\in V$.
The edge information is stored in a $m\times\left|e\right|$ matrix, where $m$ represents the number of edges and $\left|e\right|$ represents the number of edge attributes.
The graph topology information, which nodes $i,j$ are connected by which edge $e_{ij}\in E$, is stored in a $2\times m$ matrix in the COO\footnote{Coordinate list}-format.
\\
\newline
\textbf{Single Step Network}

In this process, the main goal is to capture the behavior of a traffic participant based on its immediate (dynamic) environment. 
A graph convolution approach is used to learn the representation of the participant's environment, where each participant is denoted by a node. 
Thus, nodes are updated depending on the outgoing edges and the corresponding neighbors.

\begin{figure}[htbp]
  \centering
  \def\svgwidth{\columnwidth}
  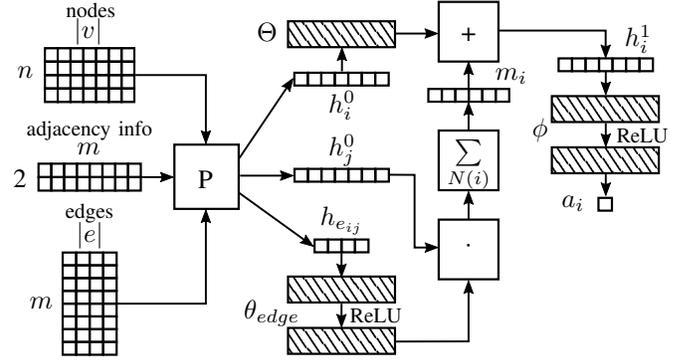 
  \caption{Graph convolution architecture}
  \label{fig:nnconv_net_architecture}
\end{figure}

In general, the graph convolution approach consists of two steps, the message step (\Cref{eq:message}) and the update step (\Cref{eq:update}). In \Cref{fig:nnconv_net_architecture}, a schematic illustration of our graph convolution architecture is depicted. Our message passing approach is based on the work of Gilmer et al.~\cite{gilmer_neural_2017}.

\begin{align}
    m_i^{s+1} = \frac{1}{\left|N(i)\right|}\sum_{j\in N(i)} h_{j}^s \cdot \theta_{edge}(e_{ij})
    \label{eq:message}
\end{align}
\begin{align}
    h_i^{s+1} = h_i^s\cdot \Theta+m_i^{s+1}
    \label{eq:update}
\end{align}
A message $m_i^{s+1}$ for each node $i$ is generated regarding the neighbor node's state $h^s_j$ and the corresponding edge attributes $e_{ij}$, for all neighbors $j\in N(i)$. 
This is done by the propagation module P (see \Cref{fig:nnconv_net_architecture}) in which all outgoing edge attributes $h_{e_{ij}}$ of the root node $i$ are extracted with the corresponding initial attributes $h^0_j$ of the neighbor node $j$.

The edge attributes of $e_{ij}$ are fed through a neural network $\theta_{edge}$, which consists of two fully connected layers and an interposed activation function. Subsequently, the current hidden state of the root node is updated by a learnable weight matrix $\Theta$. By adding this state and the message $m_i^{s+1}$ generated in the previous step, the state is updated to the new hidden state $h^{s+1}_i$.

Our network is constructed in such a way that several of these message passing steps $(0 \leq s \leq S)$ can be connected in series. This propagates the information of the individual nodes further into the graph. In this work, the GNN architecture contains only one step of message passing $(s=S)$ to demonstrate the principle function of the model and to keep it as simple as possible.

After the message passing step, each node in the graph has a hidden state depending on its neighbors and the corresponding relations. Finally, each $h_i^S$ is mapped to a single floating point number which represents the predicted acceleration $a_i$ via a neural network $\phi(\cdot)$, which consists of two fully connected layers with an interposed activation function.
\begin{align}
    a_i = \phi(h_i^S)
\end{align}

In our implementation $\theta_{edge}$ is an MLP with a hidden size of 32. Hidden states $h$ have the size of 64. $\phi$ is also an MLP with a hidden size of 128. All MLPs have an interposed ReLU activation function.

All hyperparameters were optimized manually and empirically.
\\
\newline
\textbf{Recurrent Network}

\begin{figure}[htbp]
  \centering
  \def\svgwidth{\columnwidth}
  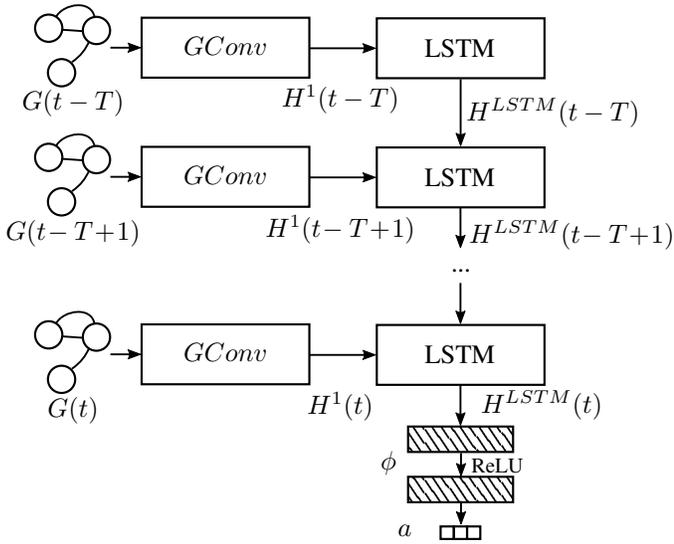 
  \caption{Recurrent graph neural network architecture}
  \label{fig:rgnn_net_architecture}
\end{figure}

In addition to considering only a single traffic scene, we examine the influence of a temporal sequence of scenes on motion prediction. To capture the temporal information, we use an LSTM architecture (see \Cref{fig:rgnn_net_architecture}). Our temporal architecture is based on the works of Taheri et al.~\cite{taheri_predictive_2019}.
 
A sequence can contain up to $T$ previous scenes. Hereby, $T$ is practically only limited by the dataset and the scenes occurring in it. It should be noted that only the acceleration of the traffic participants that are in the scene at the starting point $t=0$ are predicted. Traffic participants that are in the sequence but leave the scene earlier are still considered for the calculation of the historical states.

At first, we use the graph convolution $GConv$ architecture described above to generate a hidden state for each traffic participant ($H^1(t) = \{h^1_0(t),...,h^1_n(t)\}$) with respect to its neighbors at a given time $t$ (compare  \Cref{fig:nnconv_net_architecture} and \Cref{eq:message}, \Cref{eq:update}).
These states are fed into an LSTM block \cite{hochreiter_long_1997}, which updates the LSTM encoding $H^{LSTM}$. This step is then repeated for every scene in the sequence to continuously update the LSTM encoding.
The final encoding $H^{LSTM}(t)$ is then mapped to the corresponding acceleration values $a = \{a_0, a_1, ..., a_n\}$ for each node using $\phi$.

\section{Experiments}
\label{sec:experiments}
In this section, we evaluate the results of the two proposed network architectures. 
With the experiments, we want to show that the scene graphs can help to provide contextual information of a traffic scene in respect to behavioral prediction. 
For each entity $i$ in $G(t)$, the acceleration $a_i$ is set as the respective label. The acceleration is calculated using the ground truth trajectories' velocities $vel_i$ of entity $i$.
\begin{align}
    a^{label}_i(t) = \frac{vel_i(t+\delta_t) - vel_i(t)}{\delta_t}
    \label{eq:acceleration}
\end{align}
It should be noted that acceleration prediction is only a single use case of the proposed approach. It intends to demonstrate the applicability and to verify the functionality.
In our experiments, we set $\delta_t$ to 10 time steps, which corresponds to a time period of $1s$ in the datasets used. 
We evaluate our model, how well acceleration can be imitated using two different motion datasets.

We compare our model against a set of simple baselines for two reasons: First, we want to measure the benefit of a spatial semantic scene graph and second, to the best of our knowledge, there are no comparable approaches based on our input data.
Many recent approaches use a rasterized bird's-eye image of the scenes~\cite{bansal_chauffeurnet_2018,Chai2019MultiPathMP} or at least the coordinates of traffic participants as input. 
The work of Ma et al.~\cite{ma_multi-agent_2021} is still the closest related to our approach, but it also uses the image information for the edge attributes.
Therefore, two straightforward baselines were used:
Baseline \emph{Mean} emulates a model that always outputs the mean value over the whole test dataset. Baseline \emph{Zero} emulates a model that always outputs zero as the acceleration value, regardless of the input.

\subsection{Training}
We trained our model using Adam~\cite{kingma2014adam} with an initial learning rate of $10^{-3}$ and a batch size of 1. As a learning rate scheduler, we employed the \emph{ReduceLROnPlateau} method with a factor of 0.1 and a patience value of 10. 
The number of epochs is set to a maximum of 200 with an early stopping mechanism after 25 epochs without validation improvement. In our experiments, the trainings stopped at roughly 75 epochs, depending on the model and dataset. Furthermore, we clipped the gradients to a global gradient norm of 1.

Our model is trained on the INTERACTION dataset with 13603 graphs, corresponding to 59972 entities.
The model is subsequently tested with 11871 unseen graphs or behavior of 57413 traffic participants.
The training set of PandaSet dataset consists of 4977 graphs and the testing set consists of 2923 graphs.

During the training phase, we used the loss defined in \Cref{eq:loss_all} for the INTERACTION dataset and the loss defined in \Cref{eq:loss_pandaset} for the PandaSet dataset.

\begin{align}
    Loss = \frac{1}{n}\sum_{i\in V}\left| a_i - a^{label}_i \right|
    \label{eq:loss_all}
\end{align}

\begin{align}
    Loss_{pandaset} = \left| a_0 - a^{label}_0 \right|
    \label{eq:loss_pandaset}
\end{align}

Due to the inaccurate vehicle tracking described in \Cref{sec:preprocessing}, only the acceleration of the measurement vehicle $a_0$ is predicted for the PandaSet dataset and also used for training.

\subsection{Results}
The results are calculated in \Cref{tab:error_table} for each model as linear error (L1-loss) or as squared error (MSE-loss).
Compared to the baseline, the Single Step model (see \Cref{fig:nnconv_net_architecture}) is about 20\% better at predicting accelerations. The Single Step model, with no edge data used, serves as an ablation. Only the node information is fed into the neural network, and all edge information is set to zero.
Compared to the ablation model, our model performed about 12\% better. 
The temporal network architecture performs significantly better. Compared to the baseline, better results are achieved between 70\% and 73\%, depending on the length of the sequence under consideration. Sequences with 5, 10 and 15 scenes are considered. The longer the time history, the better the acceleration of the individual traffic participants can be predicted. However, in our experiments there are only minimal changes in the errors, when considering 10 or 15 scenes per sequence.

To make the results comparable, we assume the calculated acceleration to be constant for future time steps and calculate a trajectory based on the ground truth path of a traffic participant.
This would describe the deviation of the driven distance, which is comparable to the \emph{Final Displacement Error} (FDE) of a trajectory comparison, as follows:
 \begin{align}
     \textrm{FDE}_t = \left| a_i - a^{label}_i \right|  \frac{t^2}{2} 
 \end{align}
With our Recurrent15 model of the INTERACTION dataset, this would give a $\mathrm{FDE}_3 =0.765\,m$. However, it should be kept in mind that so far our model only predicts the longitudinal motion component (accelerating/breaking) of a traffic participant and neglects the lateral component (steering).

If we compare the predictive performance of our model with those of state-of-the-art models \cite{interpret_challenge}, we perform in the lower range of common predictors.
However, there are obvious reasons for that since we optimize only for an acceleration value and not for a future trajectory, i.e., the future course of the road is not included in the model. The curvature of the road, regulating road markings, such as a stop line, is also not considered yet. 
Instead, we show that relational information in a traffic scene helps to enable more meaningful predictions in general. Our approach complements the previous work and does not intend to replace it.
\begin{table}[htbp]
    \centering
    \begin{tabular}{lccc}
    \hline
    Model  & Dataset & L1 & MSE \\ \hline
    Single Step & INTERACTION & 0.494 & 0.451 \\
    Single Step no edge data & INTERACTION & 0.552 & 0.538 \\
    Recurrent5 & INTERACTION & 0.271 & 0.14 \\
    Recurrent10 & INTERACTION & 0.188 & 0.098 \\
    Recurrent15 & INTERACTION & \textbf{0.170} & \textbf{0.085} \\
    Baseline Mean & INTERACTION & 0.654 & 0.740\\
    Baseline Zero & INTERACTION & 0.599 & 0.622\\
    \\
    Single Step & PandaSet & 0.332 & 0.378 \\
    Recurrent5 & PandaSet & 0.259 & 0.294 \\
    Recurrent10 & PandaSet & 0.178 & 0.241 \\
    Recurrent15 & PandaSet & \textbf{0.158} &\textbf{ 0.206} \\
    Baseline Mean & PandaSet & 0.347 & 0.489 \\
    Baseline Zero & PandaSet & 0.34 & 0.489 \\
    \hline
    \end{tabular}
    \caption{Evaluation results of the tested models}
    \label{tab:error_table}
\end{table}

The relative trend of the prediction models between the INTERACTION and the PandaSet dataset is similar (see \Cref{tab:error_table}). However, comparatively inferior results are achieved with the PandaSet.
One possible explanation is that the PandaSet dataset contains about 20 times less training samples than the INTERACTION dataset because in the latter, a future acceleration is estimated for each traffic participant and not only for the ego vehicle. In addition, the PandaSet contains significantly larger numbers of participants per traffic scenes. On average, it contains about 24 traffic participants, each with 12 relations to their neighbors. In the INTERACTION dataset, there are on average about 5 vehicles in the scene and each object has about 2 relations. This reduces noise and allows the model to focus more precisely on individual traffic participants.

\section{Conclusion}
\label{sec:conclusion}
In this work, we present two models that are capable of predicting the movement of traffic participants based on spatial semantic scene graphs. 
By means of the scene graphs, semantic relationships among traffic participants are encoded. 
Through the ablative analysis, it is shown that these play an important role in the behavior of the traffic participants. In addition, the temporal model empirically shows that the preceding scenes provide important contextual information for the prediction task, and lead to significantly better results. With the inclusion of previous scenes, up to 46\% better results are achieved for the PandaSet and up to 73\% better results are achieved for the INTERACTION dataset compared to the baseline.

In the future, more semantic relation types will be investigated. While the presented model is able to represent various types of traffic participants and their different types of relations, e.g., pedestrians are not included yet, and only spatial relations are considered in our experiments. In the upcoming work, the influence of other traffic participants will be investigated. In addition, the used scene graph is limited to a few attributes, which will be extended in the future by further semantic information, possibly from external sources. 

Moreover, we will also investigate to what extent the contextual information of the scene extracted by the presented model can be used to improve a state-of-the-art image-based trajectory predictor.

\bibliographystyle{IEEEtran}
\bibliography{references}

\end{document}